\documentclass[runningheads]{llncs}
\usepackage{subfig}

\usepackage[T1]{fontenc}
\usepackage{graphicx}
\begin{document}
\title{Connected Dependability Cage: Run-Time Function and Anomaly Monitoring for the Development and Operation of Safe Automated Vehicles}
\titlerunning{Connected Dependability Cage: Run-Time Monitoring for Safe AV}
%

\author{Iqra Aslam \and
Nour Habib \and
Abhishek Buragohain \and
Meng Zhang  \and Andreas Rausch  \and Vaibhav Tiwari  \and Mohamed Benchat
}
\authorrunning{I. Aslam et al.}
%
\institute{Institute for Software and System Engineering, Clausthal University of Technology, Germany
\email{\{iqra.aslam, nour.habib, abhishek.buragohain, meng.zhang, andreas.rausch, vaibhav.tiwari, mohamed.benchat\}@tu-clausthal.de}}
\maketitle              
\begin{abstract}

The advancement of automated vehicles introduces complex safety challenges, particularly in dynamic and unpredictable environments where AI-enabled perception systems must operate reliably. Ensuring compliance with safety standards such as ISO 26262 and ISO/PAS 21448 (SOTIF) is essential for addressing system malfunctions and mitigating unsafe behavior in unknown scenarios. However, as automation levels increase, vehicles must go beyond conventional functional safety by incorporating fail-operational capabilities that enable continued safe operation during system or component failures and the handling of unfamiliar or degraded operational conditions. To address these safety concerns, we propose the Connected Dependability Cage, an architectural framework designed to enable hierarchical fail-operational behavior in AI-enabled perception systems. This framework integrates two complementary monitoring mechanisms: a Function Monitor that oversees multiple heterogeneous AI-based perception pipelines and detects inconsistencies through a voting mechanism, and an Anomaly Monitor that evaluates the reliability of AI perception by detecting unknown or novel objects in scenes that may be excluded from the training dataset. In the presence of critical discrepancies, the system supports graceful degradation, ultimately enabling a transition to a minimal-risk maneuver strategy. Furthermore, whenever either monitor raises a safety flag, an automated data recording process is initiated to facilitate iterative system development and continuous improvement. Both monitors have been implemented and validated through extensive vehicle testing, demonstrating their practical effectiveness in real-world applications.

\end{abstract}
\section{Introduction} \label{introduction}
As the level of vehicle automation increases, particularly beyond SAE Level 3 \cite{sae2018taxonomy}, the responsibility for safety and control shifts significantly towards the automated system. At the highest levels (SAE Levels 4 and 5), safety standards like ISO 26262  \cite{iso26262}  mandate fail-operational capabilities, requiring continued safe operation even in the presence of component failures. However, this growing autonomy also introduces new challenges, i.e., Safety of the Intended Functionality (SOTIF), as highlighted in ISO 21448 \cite{iso21448}. These challenges include dealing with unknown or unforeseen scenarios and ensuring safe behavior in complex, highly dynamic environments, especially for AI-enabled perception systems. To ensure compliance with applicable safety standards, the following three key challenges must be addressed:

\textbf{(a) Constructing safety arguments for AI-enabled perception systems:} Conventional safety standards such as ISO 26262 and ISO 21448 presuppose the availability of exhaustive, well-defined requirements specifications. However, such specifications are often not feasible for AI-enabled perception systems, which operate based on learned statistical correlations derived from large datasets rather than deterministic logic. While high-level functional goals—such as object classification—can be defined, the translation into low-level specifications (e.g., pixel-level behavior) remains a major open issue. For instance, pedestrian detection must generalize across a wide range of visual conditions and human postures at the pixel level—variations that are inherently difficult to formalize in explicit specifications. Despite these challenges, the automotive industry currently lacks standardized procedures to justify and argue the safety and correctness of AI-based perception systems. Therefore, robust and transparent methodologies are urgently needed to validate the performance and justify the deployment of such systems in safety-critical domains like automated driving.

\textbf{(b) Ensuring reliable operation in the presence of failures:} To operate safely, automated vehicles must consistently interpret their surroundings and make correct decisions in dynamic, hardly predictable environments. Central to this is situational awareness, where perception systems mus correctly distinguish between objects—e.g., recognizing the difference between a real car and a billboard image of a car. Misinterpretations can lead to unsafe decisions. For instance, the static billboard misclassified as a dynamic object may trigger unnecessary maneuvers. At SAE Levels 4 und 5, vehicles are required to remain safe even under systems failures, which may involve continued operation or a transition to a safe state (e.g., minimal risk condition). To address this, our architecture introduces a redundant, independent perception path, potentially based on non-AI methods. However, redundancy alone improves fault tolerance but does not eliminate the core problem: the automated systems must determine which perception path accurately reflects the surrounding environment. This challenge is conceptually akin to Byzantine failure scenarios \cite{lamport1982byzantine}, where components may fail arbitrarily and yield conflicting outputs. To mitigate such risks, run-time monitoring, consistency checks, and appropriate decision fusion mechanisms are essential to uphold safe operation in the presence of inconsistencies or degraded performance.

\textbf{(C) Data diversity and relevance:} Higher levels of automation dramatically increase the volume of sensor data to be processed in real time. For example, Waymo’s autonomous vehicles (AVs) \cite{kalra2016driving} generate over 426 GB of data per kilometer at speeds of 24 km/h. Similarly, the sensor setup used in the SafeWahr project \cite{avl2024,safewahr2024} generates approximately 33.3 GB per kilometer per vehicle. Validation of automated driving functions—such as highway driving—may require upwards of 6.62 billion kilometers of testing \cite{wachenfeld2015freigabe}, translating to approximately 1.53 billion terabytes of data. These numbers illustrate the enormous data and validation as well as certification burden posed by real-world deployment. Given that exhaustive testing of all conceivable scenarios is infeasible, residual uncertainty about uncovered edge cases remains a major challenge. To address this, future validation strategies must prioritize the acquisition of diverse and highly relevant data, complemented by techniques to detect and handle unforeseen critical events during real-world operations.

To tackle these challenges, this paper presents a comprehensive blueprint for a hierarchical fail-operational architectural framework designed to enhance the safety of automated vehicles, with a particular focus on the integrity of AI-enabled functions like environmental perception. The proposed architecture employs continuous monitoring of perception systems via two complementary mechanisms: a Function Monitor and an Anomaly Monitor, both aimed at detecting safety risks and performance deviations at runtime. In response to detected issues, the system can trigger fail-operational transitions, perform graceful degradation through an integrated reconfiguration mechanism or trigger the minimal risk maneuver like emergency stop.

The whole paper is structured as follows: an overview of related work is described in Section II. In Section III, the proposed hierarchical architectural framework, building upon the concept of the Connected Dependability Cage is explained. Sections IV and V delve into the core components of the framework: Function Monitor and Anomaly Monitor. It describes how these components work together to ensure system-level consistency and identify unforeseen operational scenarios. Section VI presents the implementation and evaluation of the framework in a practical context. Finally, Section VII concludes the paper and highlights promising avenues for future research.

\section{Related work}
 To ensure safe operation of automated vehicles, significant research has been directed towards developing fail-operational systems for automated driving. These systems are designed to maintain safety by detecting and mitigating potential failures, allowing the vehicle to continue operating with reduced functionality or gracefully degrade in performance \cite{stolte2021taxonomy}. As noted by \cite{messnarz2019highly}, the automotive industry is progressively adopting fail-operational architectures, driven by legal and insurance considerations as responsibility transitions from human drivers to the automated vehicle's technical systems. In a related vein, \cite{li2019fail} explores fail-operational strategies for Steer-By-Wire systems, with a focus on hardware-centric solutions that substitute traditional mechanical steering with electronic control systems. These developments align with the broader demand for robust fail-operational frameworks, as discussed by \cite{messnarz2019highly}, highlighting the necessity for resilient designs capable of handling operational failures. However, these approaches predominantly target deterministic control systems and lack the integration of runtime monitoring essential for AI-enabled perception systems in automated vehicles.

Recent research has explored various avenues for ensuring the safety of autonomous vehicles through runtime monitoring. One prominent approach, introduced by \cite{castelino2024connected}, leverages connected vehicle technology to create a  reliable Common Environment Model. Their method uses data from both connected vehicles and infrastructure-based sensors (V2X) for the runtime verification of an Automated Driving System (ADS). The V2X approach is great to have as an extra layer of safety when available, making it complementary but distinct from the onboard, fail-operational frameworks required for SAE Level 4+ vehicles.

Following this, \cite{ge2022systematic} developed a system to detect unusual behavior in autonomous delivery robots during operation. However, this system mainly focuses on identifying problems with individual sensors and does not have the ability to reconfigure the overall operation of the robot when any serious issues arise. On the other hand, \cite{sherry2020autonomous} explored how to monitor self-driving shuttle buses and highlighted the importance of constantly collecting data from sensors and identifying any malfunctions to ensure safe and reliable operations. While their work emphasizes the need for strong monitoring systems, it does not delve into system-level reconfiguration required to handle major failures.

In previous research, many studies have made significant contributions to developing fail-operational architectures, focusing largely on deterministic control systems and hardware redundancy. However, they often overlook the unique challenges posed by modern, AI-enabled perception systems. There is a specific need for a fail-operational architecture that is explicitly designed to ensure the safety of the AI-enabled perception system, particularly when it encounters untrained or unknown environments where its performance is uncertain. This work addresses this gap by proposing a hierarchical architectural framework based on the Connected Dependability Cage concept \cite{aniculaesei2018towards}. The proposed framework is specifically designed to ensure fail-operational functionality for the AI-enabled perception system by integrating two distinct runtime monitors (Function and Anomaly, respectively).
Together, these monitors enable the detection of safety risks and trigger fail-operational responses, such as reconfiguration or graceful degradation for executing minimal-risk maneuvers, through the automated driving system’s configuration interfaces.
\section{Connected Dependability Cage}\label{CDC}

\begin{figure}
\centering
\includegraphics[width=\textwidth]{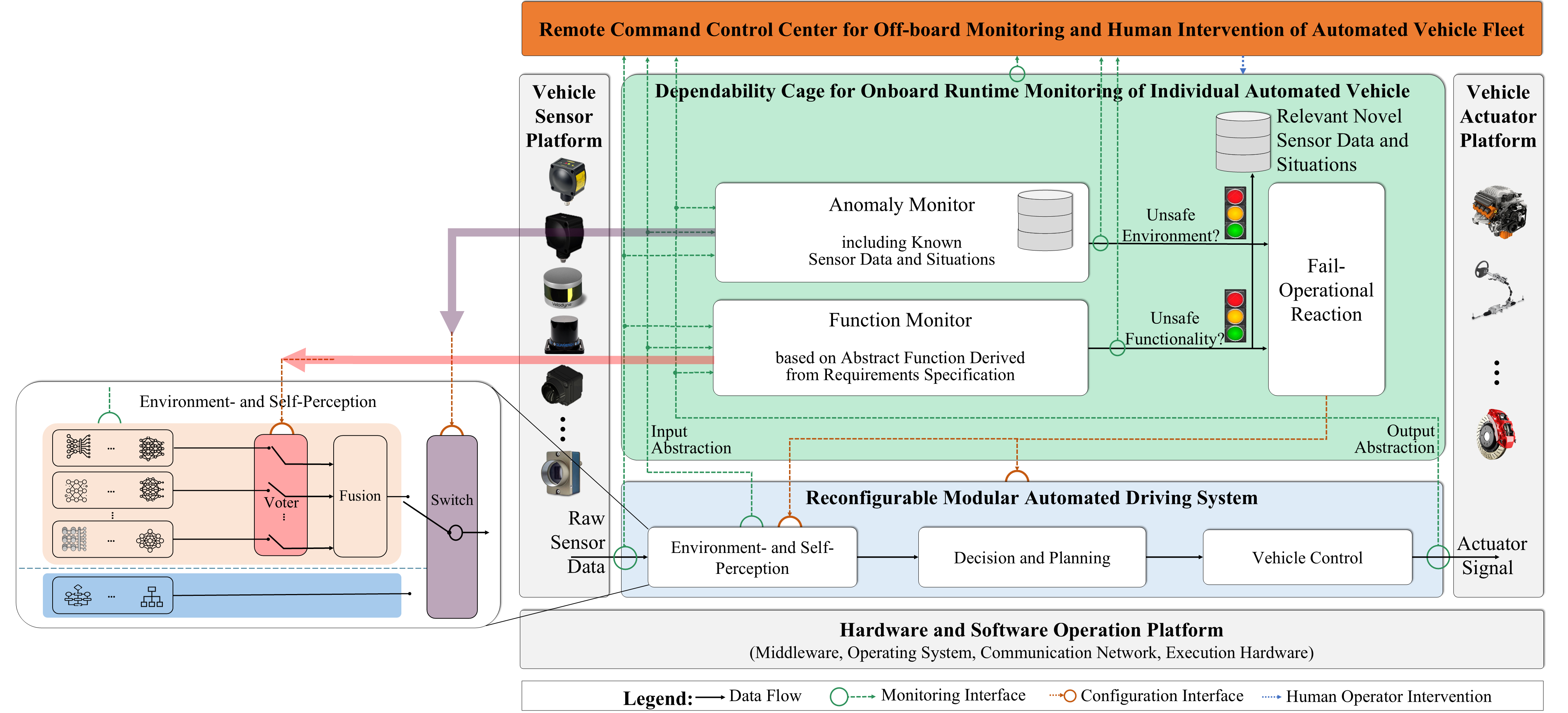}
\caption{Connected Dependability Cage using Function and Anomaly Monitors.}\label{connected_dc}
\end{figure}

The Connected Dependability Cage \cite{aniculaesei2023connecteddc,aniculaesei2025improving,aniculaesei2018towards,mauritzdiss,mauritz2015,mauritz2016,mauritz2014} is a system architecture concept that was developed to engineer dependable autonomous systems, as illustrated in Figure  \ref{connected_dc}. The system architecture comprises two monitoring subsystems with distinct roles: (1) an onboard runtime monitoring framework (highlighted in green), known as Dependability Cage which facilitates runtime monitoring of an individual AV, and (2) a remote Command Control Center (CCC, highlighted in orange) that is responsible for overseeing the fleet and enabling human operator intervention.

In addition to these subsystems, the architecture also includes the monitored system, represented by a reconfigurable modular automated driving system (highlighted in light blue). The system follows a standard high-level functional pipeline consisting of multiple modules: (1) environment and self-perception, (2) decision and planning, and (3) vehicle control \cite{behere2016,Mauer2015}. Various runtime monitors within the Dependability Cage can be developed using existing development artifacts to ensure system reliability. This monitoring framework addresses two key safety aspects: (a) validating the correctness of system behavior in accordance with dependability requirements, realized by the Function Monitor (formerly known as the qualitative monitor), and (b) ensuring the system operation through detecting unknown scenes and environments while the previous system development and testing, handled by the Anomaly Monitor (formerly the quantitative monitor). 

Both monitors require consistent and abstracted access to the monitored system. This is achieved through input and output abstraction components, as indicated by the dotted green arrows in Figure 1. These components act as standardized interfaces between the autonomous system and the monitoring modules. They utilize defined interfaces to access system data and transform it into abstract representations, whose types and values are derived from the system's requirements specification and defined dependability criteria.
The Function Monitor evaluates the system's correctness and safety under the assumption that is is operating within an environment conforming to its specification. It includes an abstract behavior function and a conformity oracle. The abstract behavior function computes, in real time, a set of actions that are correct and safe for the current abstract scene faced by the system. Meanwhile, the conformity oracle compares the output abstraction with the correct and safe abstract actions produced by the abstract behavior function. 
In contrast, the Anomaly Monitor focus on the abstract scenes encountered during runtime. It determines, in real time, whether each encountered abstract scene has been covered during development and testing, specially, by assessing whether any contained object constitutes an anomaly with respect to the known distribution of training data, without requiring manual labeling in advance.

A knowledge base provides information on previously tested scenes at an abstract level, using unique canonical representations of these abstract scenes. If the Function Monitor or Anomaly Monitor detects incorrect or unsafe behavior in the system, safety measures must be enacted to guarantee the fulfillment of dependability requirements. A fail-operational reaction component takes charge of transitioning the compromised system to a safe state with an acceptable level of risk. One possible response is the graceful degradation of the autonomous system, as described in \cite{aniculaesei2019graceful}. Additionally, when safety measures are initiated, the system automatically records data. This recorded data is invaluable for further development, as it helps identify and correct previously unknown faulty behaviors while also extending the system's test coverage.

The second subsystem in the system architecture, known as the remote CCC, complements the onboard Dependability Cage framework by enabling off-board monitoring of multiple AVs, augmented by human operator interactions. As illustrated in Figure \ref{connected_dc}, the Dependability Cage framework utilizes its fail-operational response component to maintain vehicle dependability in the face of critical issues. However, for challenges that exceed its capabilities, the CCC provides additional human intervention.

The CCC plays a crucial role in monitoring and ensuring the safety of vehicles through advanced sensory data streams, including LiDAR point clouds and camera images. When the monitoring systems identify an issue that cannot be autonomously resolved by the technical system, it triggers an alert, transferring the control and safety responsibilities to a human operator. This operator can intervene directly using the CCC's graphical user interface to modify vehicle settings like driving modes for adjusting configurations or application parameters or to initiate remote emergency stops if necessary, as detailed in  \cite{aniculaesei2023connecteddc}. In more severe scenarios, teleoperation is also available, and if necessary, the remote safety operator can dispatch local human assistance. Once the issue is resolved, the responsibility for safety and control is reverted to the vehicle's technical system, upon confirmation from the safety operator.

The Dependability Cage and remote CCC operate as distributed systems, utilizing middleware technologies to facilitate communication. This design enables the seamless transfer and sharing of safety and control responsibilities between technical systems and human operators. Until now, research has demonstrated the Connected Dependability Cage architecture across various use cases, with real-world prototypes showcasing its effectiveness \cite{aniculaesei2023connecteddc,aniculaesei2025improving,aniculaesei2018towards,mauritzdiss,mauritz2015,mauritz2016,mauritz2014}. In those studies, the monitors prioritized maintaining system-wide safety by overseeing the complete automated driving pipeline. In this paper, we refine that approach by concentrating on the environmental perception module within the automated driving system. Beyond initiating fail-safe measures like emergency braking, we introduce an architectural framework aimed at achieving hierarchical fail-operational responses within the perception system, which will be further elaborated in the following section.

\subsection{Hierarchical Fail-Operational Architectural Framework}

The environmental perception module is a crucial part of automated driving system, as it forms the foundation for how the vehicle interprets its surrounding environment. It processes raw sensor data—such as from LiDAR, radar, and cameras—into a coherent representation of the external world. Given its criticality, we propose a comprehensive blueprint for a hierarchical fail-operational architecture framework specifically designed to ensure the dependability of this module. As illustrated on the left side of Figure {\ref{connected_dc} }, the proposed framework is based on a hierarchical safety concept that incorporates redundancy, consisting of two distinct and independent perception paths:

\begin{itemize}
\item  \textbf{Primary AI-Enabled Path} (highlighted in light orange), which leverages the advanced capabilities of AI-enabled perception models.
\item  \textbf{Fallback Deterministic Path (non-AI)} (highlighted in blue), which serves as a crucial safety-critical backup, relying on simpler, non-AI algorithms to support graceful degradation. 
\end{itemize}

 Each of these perception layers is supported by a dedicated monitor—Function Monitor and Anomaly Monitor, respectively—designed to proactively detect and respond to potential hazards ({c.f Section \ref{introduction}}). This hierarchical structure not only ensures immediate operational safety but also facilitates continuous improvement of the system through iterative development and feedback.


\subsubsection{Primary AI-Enabled Path: Reconfiguration within AI-enabled environment and self-perceptions} \label{primary_reconfig} 

The primary perception path, highlighted in light orange in Figure \ref{connected_dc}, follows the principle of "safety in numbers" to avoid single points of failure. Instead of depending on a single AI model, the path incorporates multiple heterogeneous AI-enabled perception models operating concurrently. Each model processes sensor inputs (e.g., LiDAR, radar, cameras) and produces its own object list, including the objects' semantic information like classes, positions, and orientations of detected objects. 
The outputs of these models are fused into a single coherent perception output to achieve a more comprehensive understanding of the environment. However, model heterogeneity also introduces the risk of inconsistencies between outputs. 

To address this, a voter mechanism (highlighted in red in Figure \ref{connected_dc}) is implemented within the primary perception path. This voter filters out inconsistent outputs from models that exhibit significant deviations, thereby excluding them from the subsequent fusion process.

This mechanism is supported by the Function Monitor (cf. Section 4), which continuously evaluates the consistency of model outputs against predefined thresholds derived from Hazard Analysis and Risk Assessment (HARA). Upon detecting inconsistencies, the Function Monitor raises a safety flag to the fail-operational reaction component (center of Figure \ref{connected_dc}), triggering a reconfiguration of the voter and thus excluding the inconsistent perception models from downstream fusion processing. In addition, the Function Monitor initiates data recording, so that the full event—including raw sensor inputs and divergent model outputs—is logged for further analysis.

The validated outputs of the remaining models are then fused into a reliable perception output. This process forms the first level of fail-operational safety within the AI-enabled perception system.

The recorded data serve as valuable training assets, helping to create a growing library of edge cases and system weaknesses. They can be used to retrain and enhance the AI-enabled perception models, thereby strengthening the system performance over time. The internal mechanisms of the Function Monitor used for consistency validation are described in detail in Section 4.

\subsubsection{Fallback Deterministic Path: Reconfiguration between AI and non-AI (deterministic) environment and self-perceptions }\label{secondary_reconfig} 

As illustrated in blue in Figure \ref{connected_dc}, the fallback deterministic perception path provides another critical layer of conceptual independence from the AI-based system. This path is based on deterministic, rule-based logic using predictable and verifiable algorithms. In this study, for example, this path is implemented via geometry-based object detection.

Rather than learning from data, this approach applies hardcoded size constraints to LiDAR point clouds, which are clustered using algorithms such as DBSCAN or Euclidean-distance-based clustering. These clusters are then analyzed using geometric and mathematical models (e.g., rule-based shape fitting) to detect and classify objects—e.g., rectangles representing vehicles and cylinders representing pedestrians.

Since this pipeline does not rely on AI, it can maintain safe system functionality even when the primary AI-enabled path fails—such as in unknown or novel scenes not represented during AI model training.

A dedicated switch component facilitates safe transitions between the AI-based and deterministic paths. This switch operates under the support of the Anomaly Monitor, which continuously evaluates whether the AI models are operating within the scenes and environments covered during development.

When the Anomaly Monitor detects a scene containing anomaly objects—i.e., outside the training data distribution—it raises a safety flag to the fail-operational reaction component, which triggers a controlled transition to the deterministic fallback perception path. This is necessary since he AI-enabled perception models, in the presence of untrained anomalies, may produce unreliable outputs. In the case of detected anomaly scenes, the Anomaly Monitor also initiates data recording, storing the anomaly-related sensor data and perception results for future retraining and system enhancement.

This mechanism enables graceful degradation: while the vehicle may lose access to its most advanced perception capabilities, it retains a safe, core awareness of its environment. This allows for continued safe operation under restricted conditions or execution of a minimal-risk maneuver, such as coming to a controlled stop.

As with the primary path, detected anomalies are systematically recorded and used to expand and improve the training datasets—thus reducing future occurrences of unseen edge cases and contributing to the ongoing maturation of the AI-based perception system.

\section{Function Monitor}\label{FM} 

As previously mentioned, the Function Monitor has been used to assess the functional correctness of entire automated driving systems. In the context of our research work, its primary role is to continuously evaluate the correctness and safety of the AI-enabled perception system. This is accomplished by comparing the outputs of multiple heterogeneous AI perception models—specifically, their generated object lists, which contain semantic attributes such as class, position, and orientation—against predefined thresholds derived from a HARA.

However, directly comparing every object in the output lists is impractical. The perception models are based on different sensor modalities, each with varying ranges and fields of view (e.g., LiDAR generally offers longer range and a wider field of view than cameras). Moreover, many detected objects are not immediately safety-critical, such as those located far from the ego-vehicle or in adjacent lanes. Comparing all objects indiscriminately would also result in frequent, non-critical mismatches, leading to unnecessary safety alerts and increased computational overhead. Consequently, the comparison must be restricted to safety-critical objects within a shared region of interest (ROI) to ensure both relevance and efficiency.

To address this challenge, the proposed approach centers on a dynamically computed ROI. By focusing the evaluation within this ROI, the Function Monitor enforces a precise and meaningful safety requirement: each perceived object within the defined ROI that is forwarded to the subsequent fusion process must be confirmed by a majority of independent, AI-enabled perception paths.

This requirement encompasses two key aspects: first, the ROI must be spatially defined around the automated vehicle; second, the objects identified within this ROI by the heterogeneous AI-based perception models must exhibit mutual consistency. To operationalize this requirement, the Function Monitor is implemented with two specialized components that directly correspond to its theoretical underpinnings. The abstract behavior function (previously introduced in high-level concept of Function Monitor, cf. Section 3) is realized through the Safe Zone, which defines the ROI, while the conformity oracle is embodied by the AI Perception Validator, which assesses the consistency of detected objects across different AI-enabled perception models.

\begin{figure}[htbp]
\centerline{\includegraphics[width=0.97\textwidth]{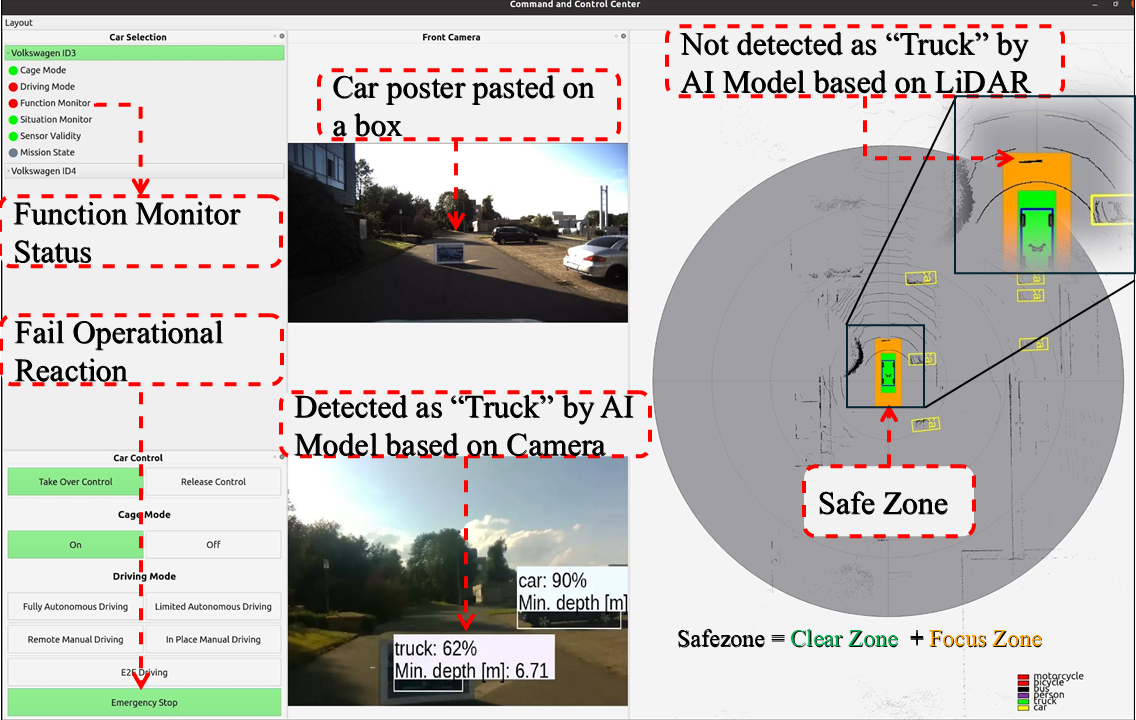}}
\caption{Function Monitor evaluation scenario.}
\label{fm}
\end{figure}
\textbf{(a) Safe Zone} is a dynamic safety area surrounding the automated vehicle, which is calculated based on the vehicle's current speed, steering angle, braking distance, and physical dimensions. Safe Zone expands around the vehicle based on the vehicle's driving direction, speed, and steering angle \cite{grieser2020assuring}. While this concept has been previously used to monitor entire automated driving systems (e.g., by ensuring it is free from obstacles), its role in this work is more specific: to serve as a dynamic filter for validating the AI-enabled perception models.

As defined in prior work \cite{helsch2022qualitative}, the Safe Zone is designed to support different operational modes of the vehicle and is divided into two sub-zones \cite{helsch2022qualitative}, as illustrated in Figure \ref{fm}. The Clear Zone (green) is a calculated safety area based on the vehicle‘s motion for fulfilling the functional safety Requirement to prevent Collisions, while the Focus Zone (orange) is the Safety Area with tolerance for extending the Clear Zone. The Focus Zone defines the ROI when the vehicle operates with full autonomous functionality, while the Clear Zone (green) is used during degraded modes of operation with alternative configurations, such as reduced maximal speed. In the scope of this work, we do not differentiate between operational modes; instead, the entire calculated Safe Zone is used as the validation region for assessing the consistency of AI-enabled perception models. This Safe Zone thus establishes a shared spatial reference and acts as a semantic filter, excluding non-safety-critical objects and enabling the Function Monitor to focus exclusively on immediate potential hazards.






\textbf{(b) AI Perception Validator} receives as input the calculated Safe Zone and the object lists generated by multiple, heterogeneous AI-enabled perception models. It begins by filtering out any objects detected outside the Safe Zone, thereby limiting the evaluation to safety-relevant regions. For the remaining objects within this zone, the component conducts detailed, object-level attribute comparisons—matching parameters such as object class, distance, and bounding box dimensions \cite{ai_fm_ab}.

If the differences between corresponding object attributes fall within predefined thresholds derived from the HARA, the models are deemed mutually consistent, and no safety flag is raised. In this case, the validated data is considered reliable and can be forwarded to the voter component for subsequent fusion. If, however, the attribute deviations exceed the HARA-defined tolerances, the Function Monitor raises a safety flag, indicating a critical inconsistency among the AI-enabled perception models.

Through this process, the AI Perception Validator enables a majority-based consistency check among the models, ensuring that only reliable outputs are used for fusion. Inconsistent perception results are excluded from the fusion process. Furthermore, all detected discrepancies are systematically logged, providing valuable feedback for future development. These logs serve as training assets for retraining and continuously improving the performance of the AI-enabled perception models over time.


\section{Anomaly Monitor}\label{AM}
The Anomaly Monitor plays a pivotal role within the Connected Dependability Cage framework, acting as a critical safety mechanism in automated driving systems. Its primary function is to evaluate whether the system’s real-time input data remains within the distribution of data used during the AI model’s initial training. This assessment is essential to ensure the reliability and validity of the AI-enabled perception output. Positioned as a mediator, the Anomaly Monitor enables a seamless and controlled transition from the AI-enabled perception paths to the deterministic, non-AI fallback path. This transition supports fail-operational behavior when the system encounters unforeseen scenes not represented during development.

At its core, the Anomaly Monitor is designed to distinguish between known (in-distribution) and unknown (out-of-distribution) input scenes. To achieve this, it processes raw sensor inputs—such as camera images—in real time, identifying deviations from expected patterns. Advanced deep learning methods such as autoencoders \cite{rausch2021autoencoder} and GAN-based models like GANomaly \cite{habib2023towards} are employed to detect subtle irregularities in the input data that may indicate unfamiliar objects or behaviors. For example, while the class "bicycle" may be part of the training data, an unusual behavior such as a cyclist performing a wheelie—unseen during training—would be flagged as an anomaly. In such cases, the Anomaly Monitor triggers a fail-operational response by switching to the deterministic perception path to maintain safe system behavior.

Beyond real-time anomaly detection, the Anomaly Monitor also supports long-term system improvement. When an anomaly is detected and the fallback path is activated, the corresponding sensor data is automatically recorded. These novel inputs are integrated into the retraining pipeline, thereby extending and refining the training dataset. Over time, this iterative learning process enhances the performance of the AI-enabled perception system, systematically converting previously unknown scenes into known, manageable ones. This not only increases the precision of perception but also significantly improves the safety and reliability of automated driving systems in complex, real-world environments.

\begin{figure}[htbp]
\centerline{\includegraphics[width=0.97\textwidth]{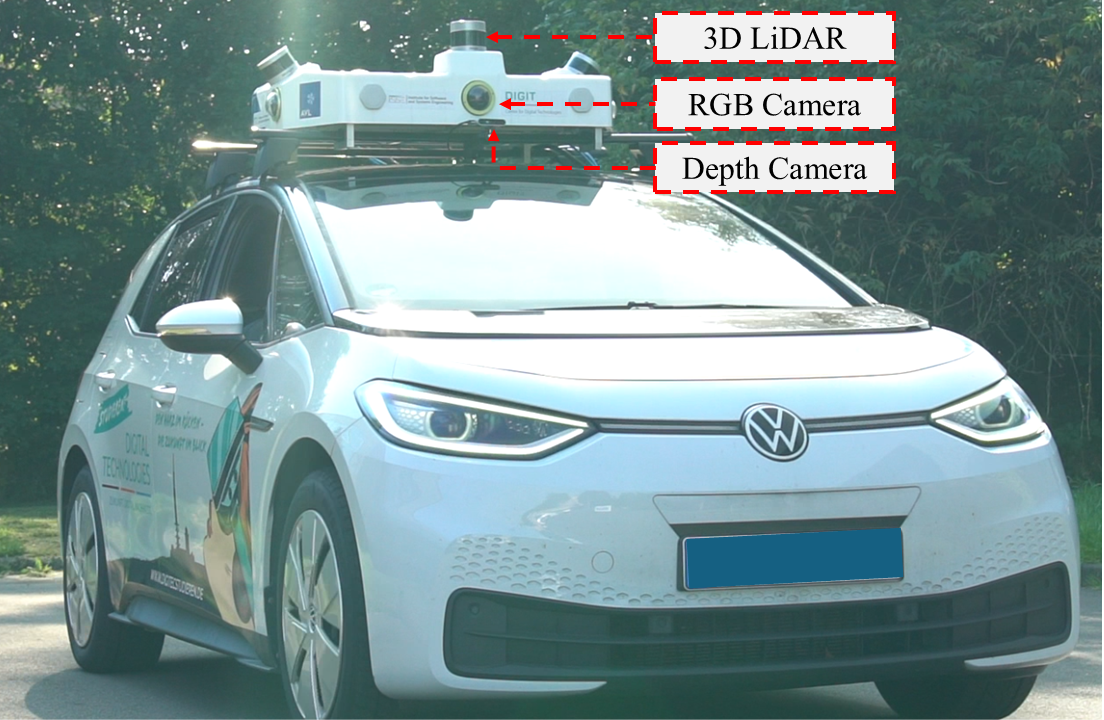}}
\caption{Test vehicle (ID3) with integrated sensor suit box (Rooftop Box).}
\label{id3}
\end{figure}

\section{Implementation and Evaluation}

This section presents the implementation and evaluation of the Function Monitor and Anomaly Monitor, as well as the fail-operational system responses triggered by these monitoring components. The evaluation also highlights the integrated data recording functionality, which is activated when the Anomaly Monitor detects unknown scenes which have not been represented during the AI model’s development. This mechanism enables the continuous collection of novel driving scenes, which are subsequently fed back into the development cycle to expand the training dataset and enhance the perception system’s ability to handle previously untrained scenes.

For the evaluation setup, a Volkswagen ID.3 vehicle was equipped with an integrated sensor platform known as the AVL Rooftop Box \cite{avl_ab} (cf. Figure \ref{id3}). The Rooftop Box features three 3D Velodyne LiDAR sensors and four RGB cameras, providing a full 360° view of the vehicle’s surroundings. Additionally, it includes a front-facing Intel depth camera \cite{intel_ab} and a GNSS sensor, which is used to calculate the vehicle’s speed and steering angle. For this initial evaluation, only the front-facing 3D LiDAR, RGB camera, depth camera, and GNSS sensor were utilized.

All sensors were calibrated relative to the center of the vehicle’s rear axle to ensure consistent spatial alignment. This calibration is essential for ensuring that the AI-enabled perception systems generate object lists within a common coordinate frame, thereby enabling accurate cross-model comparisons.

For the Function Monitor, two heterogeneous AI perception models were deployed on the testing vehicle: one based on RGB camera data (referred to as the camera-based AI model) and the other on 3D LiDAR data (referred to as the LiDAR-based AI model), using open-source implementations \cite{mmdetection3d_ab,pointpillars_ab,yolonano_ab}. For the Anomaly Monitor, anomaly detection in this initial evaluation was performed solely using data from the RGB camera.

\begin{figure}[htbp]
    \centering{\includegraphics[width=0.98\textwidth]{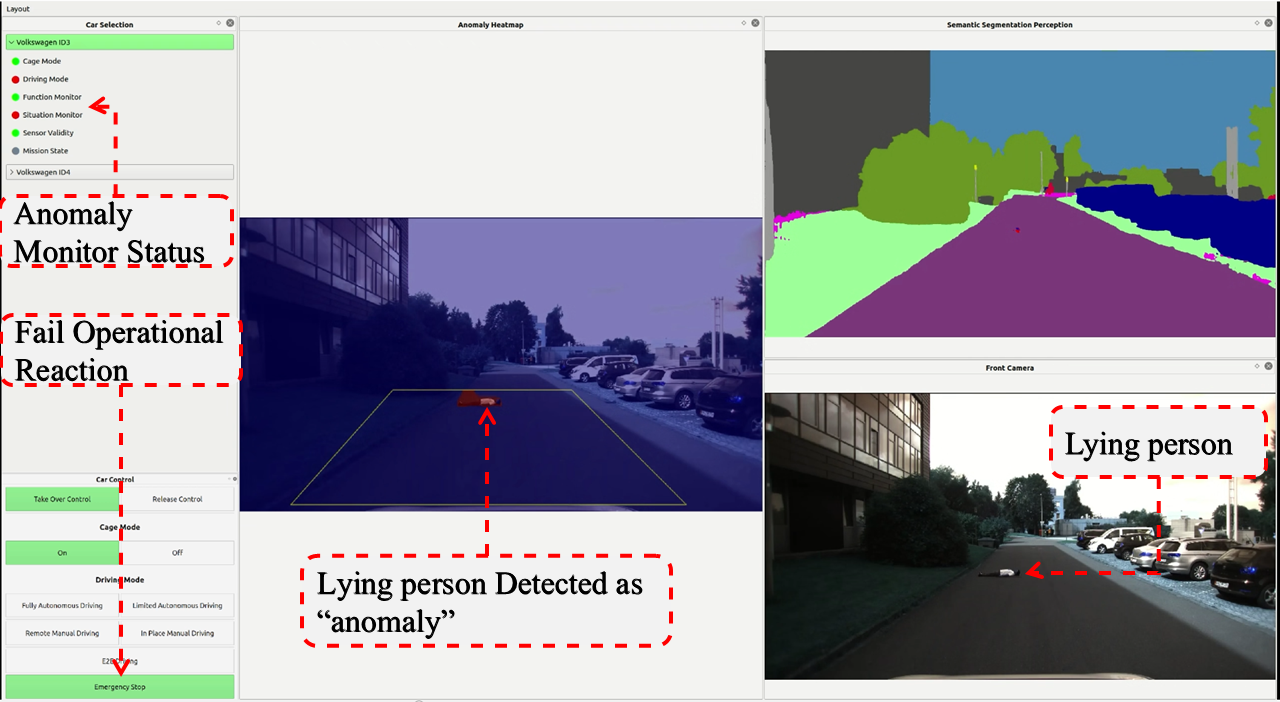}}
    \caption{Anomaly Monitor evaluation scenario: unknown behavior of known class.}
    \label{sm_1}
\end{figure}

Two test scenarios were selected to qualitatively evaluate the Function Monitor and Anomaly Monitor, utilizing the remote CCC user interface \cite{aniculaesei2023connecteddc,aslam2024runtime}. This interface provides real-time access to live sensor data streams, including LiDAR and RGB camera feeds, AI-predicted perception outputs from both modalities, anomaly heatmaps, and semantic segmentation results, as illustrated in Figures \ref{fm}, \ref{sm_1}, and \ref{sm_2}.

The test environment was a parking lot located at Arnold-Sommerfeld-Straße 1, 38678 Clausthal-Zellerfeld. To emulate realistic driving scenarios, additional static and dynamic objects were placed along the test vehicle’s path, including a pedestrian (represented by a human actor) and a car poster mounted on a box to emulate another vehicle. These objects were arranged to create diverse perception conditions for both monitors, as shown in Figures~\ref{fm}, \ref{sm_1}, and \ref{sm_2}.

\subsubsection{Function Monitor} \label{fmtext}
Figure \ref{fm} illustrates the evaluation scenario for the Function Monitor. In this setup, a box with a car poster attached is positioned on the road, as shown in the top-middle section of the figure. The camera-based AI model classifies the poster as a truck (bottom middle), whereas the LiDAR-based AI model does not associate it with any vehicle class. Instead, the LiDAR system interprets the poster merely as sparse point cloud data without assigning a semantic label (top right).

This discrepancy results in inconsistent outputs between the two heterogeneous AI-enabled perception models. As the Volkswagen ID.3 approaches the object and the poster enters the dynamically calculated Safe Zone, the Function Monitor detects the inconsistency and raises a safety flag (top left). This triggers an emergency stop as part of the fail-operational response.

This scenario demonstrates the Function Monitor’s critical role in identifying perception mismatches across redundant sensing modalities, enabling timely intervention when outputs are ambiguous or conflicting—thereby contributing to the overall safety and robustness of the automated driving system.

\subsubsection{Anomaly Monitor}
\begin{figure}[htbp]
    \centering
    {
        \includegraphics[width=0.98\textwidth]{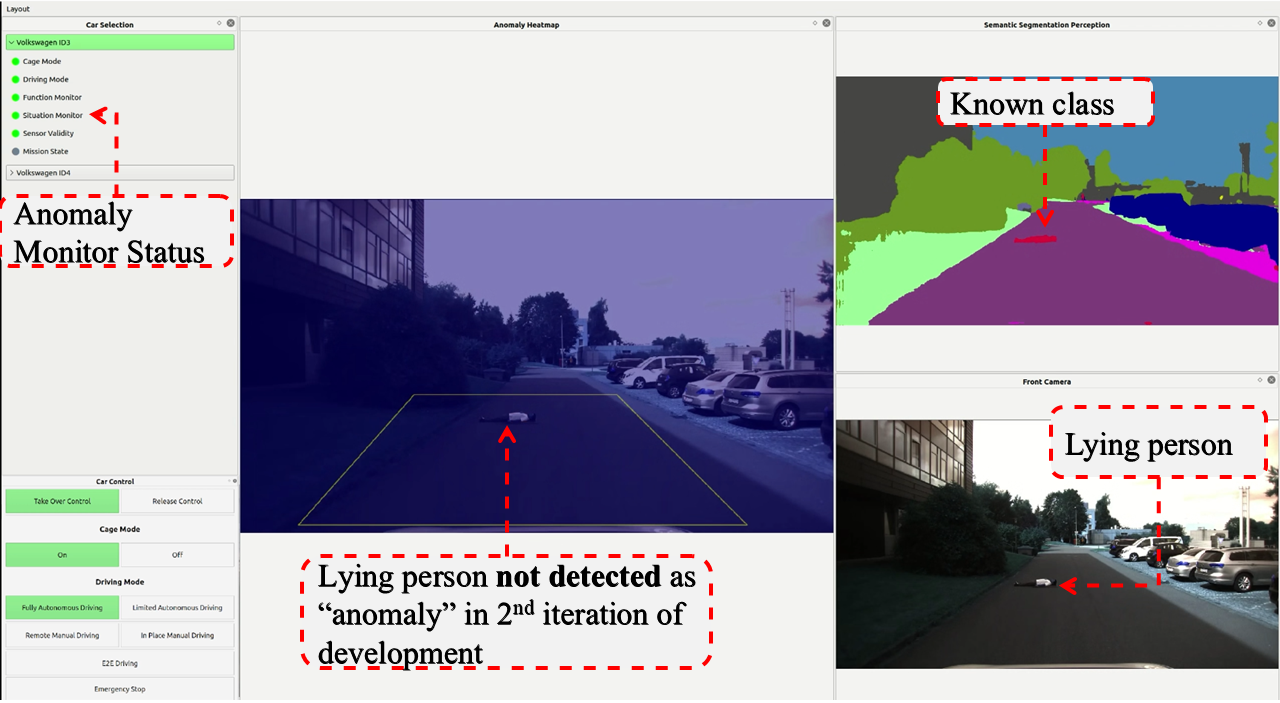}
    }
    \caption{Anomaly Monitor evaluation scenario: known behavior in the 2nd iteration.}
    \label{sm_2}
\end{figure}

Figures \ref{sm_1} and \ref{sm_2} illustrate the evaluation scenario for the Anomaly Monitor, in which a human pedestrian is lying in the middle of the road (bottom right of both figures), directly in the moving path of the Volkswagen ID.3. The evaluation consists of two phases.

In the first phase, the lying pedestrian is flagged as an anomaly (middle of Figure \ref{sm_1}). Although the object belongs to a known class (pedestrian), the unusual posture—lying on the ground—was not represented in the training data and thus was not considered during development (as discussed in Section \ref{AM}). The Anomaly Monitor successfully identifies this deviation as an anomaly, triggering both the emergency stop (top left of Figure \ref{sm_1}) and the activation of the data recording system. The recorded scene is subsequently used for retraining the AI perception model.

In the second phase, after retraining with the data captured during the first encounter, the AI model is now able to recognize the lying pedestrian as a valid instance of a known class (middle of Figure \ref{sm_2}). As a result, the Anomaly Monitor no longer flags the scene as anomalous. This is reflected in the absence of a safety flag and emergency stop in Figure \ref{sm_2}.

This scenario demonstrates the system’s ability to evolve through iterative development. By continuously incorporating previously unobserved scenarios into the training data, the AI-enabled perception system incrementally extends its coverage of real-world variations—enhancing both reliability and operational safety over time.

The initial evaluation of both monitors successfully demonstrated the core capabilities of the proposed monitoring framework. The Function Monitor effectively detected inconsistencies between heterogeneous AI-enabled perception models, while the Anomaly Monitor reliably identified previously unseen scenes by detecting either novel objects or unfamiliar behaviors of known objects that fall outside the distribution of the training data. In addition, both monitors seamlessly integrated with the data recording system, enabling the capture of such unknown instances to support iterative model refinement and enhance the long-term adaptability of the overall perception system.

It is important to note that this initial evaluation was designed as a qualitative assessment, with each monitor evaluated independently. Consequently, although the architectural framework supports a full fail-operational response, including the automated switch to a deterministic fallback path, this mechanism was not yet validated within the scope of the current study.

Building on these promising results, the next phase of our research will focus on a comprehensive quantitative evaluation, as outlined in Section \ref{CDC}. This will involve the full implementation of the fail-operational framework and rigorous testing under real-world driving conditions on public roads. The goal is to enable robust system behavior in response to both inconsistencies of perception results and unknown scenes, ensuring that the automated vehicle maintains safe operation even in complex and unpredictable environments.

\section{Conclusion and outlook}
In this paper, we presented a comprehensive blueprint for a hierarchical fail-operational architectural framework aimed at enhancing the safety of automated vehicles, with a particular focus on the integrity of AI-enabled perception functions. The framework builds on the principles of the Connected Dependability Cage approach \cite{aniculaesei2023connecteddc}, leveraging real-time Function and Anomaly Monitoring to ensure dependable and safe system behavior.

The Function Monitor is responsible for validating the behavioral correctness of the perception system by assessing the consistency of outputs from multiple, redundant AI-enabled perception models. Only those outputs that achieve agreement through a majority voting mechanism are propagated further in the driving pipeline, thereby increasing the reliability of the perception component.

In contrast, Anomaly Monitoring is designed to detect previously untested or unforeseen scenes—more precisely, anomalies such as unknown objects or atypical behaviors of known objects that fall outside the distribution of the training data. When such anomalies are detected, the system initiates a controlled switch to a deterministic, non-AI-enabled fallback perception path, effectively sidelining the AI-based perception in favor of a verified, safety-assured fallback. This failover mechanism is essential for maintaining operational safety in unpredictable or novel environments.

The integration of both monitors enables a hierarchical fail-operational capability, allowing the system to respond appropriately to both inconsistencies and unknown scenes. This layered safety architecture contributes significantly to the robustness and reliability of the overall perception system.

Both monitors have been successfully implemented and qualitatively evaluated through controlled vehicle testing, demonstrating their effectiveness in real-world scenarios. Building on this foundation, future work will focus on a comprehensive quantitative assessment of the full fail-operational framework through extensive testing on public roads. This next phase will investigate the interaction between the monitoring systems, with a particular emphasis on the underlying decision logic for model voting and failover switching. Furthermore, we aim to revisit the mode control component introduced in the original Connected Dependability Cage framework \cite{aniculaesei2023connecteddc}, and explore strategies for graceful degradation, ensuring the system can maintain a safe operational state even in degraded modes.

\subsubsection*{Acknowledgements} The authors gratefully acknowledge the collaboration and contributions of all project partners. This research was conducted within the framework of the project SafeWahr (in German: Sichere Freigabe und zuverlässiger Serienbetrieb durch kontinuierliches, echtzeitfähiges Monitoring der Umgebungswahrnehmung autonomer Fahrzeuge), funded by the Federal Ministry for Economic Affairs and Climate Protection (BMWK) in Germany under grant number 19A21026E.

%
%
%
%

\bibliographystyle{splncs04}  

\bibliography{sn-bibliography}

\end{document}